\documentclass[letterpaper, 10 pt, conference]{ieeeconf}  

\IEEEoverridecommandlockouts                              
\usepackage{amsmath, nccmath}
\usepackage[numbers]{natbib}
\usepackage{graphicx}
\usepackage{amsfonts}

\newcommand{\pp}[1]{\left( #1 \right)}

\newcommand{\bb}{\mathbb}

\title{\LARGE \bf
Eliminating All Bad Local Minima from Loss Landscapes\\Without Even Adding an Extra Unit
}

\author{Jascha Sohl-Dickstein, Kenji Kawaguchi
\\
\tt\small jaschasd@google.com, kawaguch@mit.edu}

\begin{document}

\maketitle
\thispagestyle{empty}
\pagestyle{empty}

\begin{abstract}
Recent work \citep{liang2018adding,kawaguchi2019} has noted that all bad local minima can be removed from neural network loss landscapes, by adding a single unit with a particular parameterization. 
We show that the core technique from these papers can be used to remove all bad local minima from any loss landscape, so long as the global minimum has a loss of zero. 
This procedure does not require the addition of auxiliary units, or even that the loss be associated with a neural network. 
The method of action involves all bad local minima being converted into bad (non-local) minima at infinity in terms of auxiliary parameters. 

\end{abstract}

\section{Eliminating all bad local minima}\label{sec elim}

Take a loss function $L\pp{\theta}$, with parameters $\theta$, and with a global minimum $\min_\theta L\pp{\theta}=0$. Consider the modified loss function
\begin{align}
\tilde{L}\pp{\theta, a, b} &= 
L\pp{\theta} \pp{
1 + \pp{a \exp\pp{b} - 1}^2
} + \lambda a^2
,
\label{eq aug loss}
\end{align}
where $a,b\in\bb R$ are auxiliary parameters, and $\lambda \in \bb R^+$ is a regularization hyperparameter. 
The specific form of Equation \ref{eq aug loss} was chosen to emphasize the similarity to the approach in \citet{liang2018adding} and \citet{kawaguchi2019}, but without involving auxiliary units.

As can be seen by inspection, the gradient with respect to the auxiliary parameters $a,b$ is only zero for finite $b$ when $L\pp{\theta} = 0$ and $a=0$. Otherwise, $a$ will tend to shrink towards zero to satisfy the regularizer, $b$ will tend to grow towards infinity so that $a \exp\pp{b}$ can remain approximately 1, and no fixed point will be achieved for finite $b$. 
Thus, all non-global local minima of $L\pp{\theta}$ are transformed into minima at $b \rightarrow \infty$ of $\tilde{L}\pp{\theta, a, b}$. Recall that minima at infinity do not qualify as local minima in $\mathbb{R}^n$. 
Therefore, any local minimum of $\tilde{L}\pp{\theta, a, b}$ is a global minimum of $L\pp{\theta}$, and $\tilde{L}\pp{\theta, a, b}$ has no bad local minima.

\begin{figure}
\includegraphics[width=\linewidth]{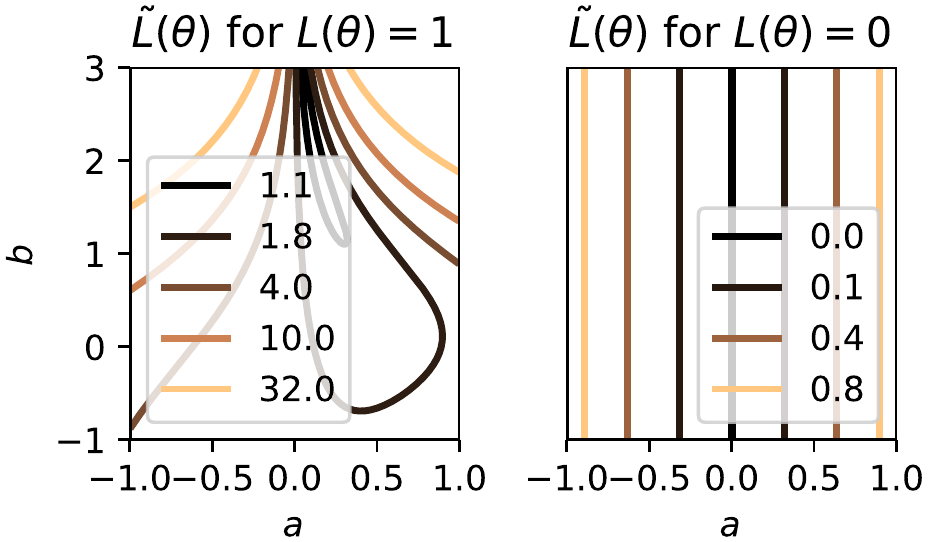}
\caption{
\textbf{All local minima of $L\pp{\theta}$ with $L\pp{\theta} > 0$ become non-local minima at infinity of $\tilde{L}\pp{\theta, a, b}$.} 
Contour plots of the modified loss landscape $\tilde{L}\pp{\theta, a, b}$ in terms of auxiliary parameters $a$ and $b$, for $\lambda=1$. When the original loss function $L\pp{\theta} > 0$, then $\tilde{L}\pp{\theta, a, b}$ approaches a minimum in terms of $a$ and $b$ as $b \rightarrow\infty$ and $a \rightarrow 0$. 
When $L\pp{\theta}$ is at its global minimum, $L\pp{\theta} = 0$, then $\tilde{L}\pp{\theta, a, b}$ has a local minimum at $a=0$, for any value of $b$.
\label{fig viz}}
\end{figure}

See Appendix \ref{sec app} for a more formal derivation, and Figure \ref{fig viz} for a visualization.

\section{Is this significant?}

By eliminating the auxiliary neurons which play a central role in \citet{kawaguchi2019} and \citet{liang2018adding} we hope to provide more clarity into the mechanism by which bad local minima are removed from the augmented loss. 
We leave it to the reader to judge whether removing local minima in this fashion is trivial, deep, or both.

We also note that there is extensive discussion in Section 5 of \citet{kawaguchi2019} of situations in which their auxiliary variable $b$ (which plays a qualitatively similar role to $b$ in Section \ref{sec elim} above) diverges to infinity. So, it {\em has} been previously observed that pathologies can continue to exist in loss landscapes modified in a fashion similar to above.

\section*{Acknowledgments}

We thank Leslie Kaelbling, Andrey Zhmoginov, and Hossein Mobahi for feedback on a draft of the manuscript.

\bibliographystyle{plainnat}
\bibliography{main}

\begin{appendices}

\section{All critical points of $\tilde{L}$ are global minima of $L$}\label{sec app}

At critical points of $\tilde{L}(\theta,a,b)$, $\partial_a \tilde L(\theta,a,b)=2L(\theta)(a \exp(b)-1)\exp(b)+2\lambda a=0$ and $\partial_b \tilde L(\theta,a,b)=2L(\theta)(a \exp(b)-1)a\exp(b)=0$, which together imply that  $a=0$. Substituting in $a=0$, we must have $\partial_a \tilde L(\theta,a,b)=-2L(\theta)\exp(b)=0$ at any critical point $(a,b) \in \mathbb{R}^{2}$ of $\tilde L$ with respect to $(a,b)$. 
This can only be satisfied by $L(\theta)=0$. 
Therefore at every critical point $\theta^*$ of  $\tilde L$ (including every local minimum),  $L(\theta^*)=0$, and thus $\theta^*$ is a global minimum of $L\pp{\theta}$.  

\end{appendices}
\end{document}